\documentclass{IOS-Book-Article}

\usepackage{mathptmx}
\usepackage{soul}\setuldepth{article}
\usepackage[numbers]{natbib}
\usepackage[fleqn]{amsmath}
\usepackage{graphicx}
\usepackage{acronym}
\usepackage{subcaption}
\acrodef{AM}{Activation Maximization}
\acrodef{ANN}{Artificial Neural Network}
\acrodef{ASR}{Automatic Speech Recognition}
\acrodef{ASG}{Auto Segmentation Criterion}
\acrodef{AUC}{area under the curve}
\acrodef{CCA}{Canonical Correlation Analysis}
\acrodef{CNN}{Convolutional Neural Network}
\acrodef{DL}{Deep Learning}
\acrodef{DNN}{Deep Neural Network}
\acrodef{EEG}{Electroencephalography}
\acrodef{ERP}{Event-Related Potential}
\acrodef{GAN}{Generative Adversarial Network}
\acrodef{Grad-CAM}{Gradient-weighted Class Activation Mapping}
\acrodef{LRP}{layer-wise relevance propagation}
\acrodef{LSTM}{Long Short-Term Memory}
\acrodef{ML}{Machine Learning}
\acrodef{MLP}{Multi-Layer Perceptron}
\acrodef{MSE}{Mean Squared Error}
\acrodef{NAP}{Neuron Activation Profile}
\acrodef{NLP}{Natural Language Processing}
\acrodef{NAvAI}{Normalized Averaging of Aligned Inputs}
\acrodef{PCA}{Principal Component Analysis}
\acrodef{PSO}{Particle Swarm Optimization}
\acrodef{ReLU}{Rectified Linear Unit}
\acrodef{RNN}{Recurrrent Neural Network}
\acrodef{SNAP}{Saliency-Adjusted Neuron Activation Profile}
\acrodef{SOM}{Self-Organizing Map}
\acrodef{SVD}{Singular Value Decomposition}
\acrodef{tSNE}{t-Distributed Stochastic Neighbor Embedding}
\acrodef{UMAP}{Uniform Manifold Approximation and Projection}

\def\hb{\hbox to 11.5 cm{}}

\begin{document}

\pagestyle{headings}
\def\thepage{}
\begin{frontmatter}              % The preamble begins here.

%\pretitle{Pretitle}
\title{Visualizing Deep Neural Networks with Topographic Activation Maps}

\markboth{}{May 2023\hb}
%\subtitle{Subtitle}

\author[A]{\fnms{Valerie} \snm{Krug}%,\orcid{0000-0002-4729-1840}%
\thanks{Corresponding Author: Valerie Krug, valerie.krug@ovgu.de}},
\author[A,B]{\fnms{Raihan Kabir} \snm{Ratul}},
\author[A]{\fnms{Christopher} \snm{Olson}}
and
\author[A]{\fnms{Sebastian} \snm{Stober}}%\orcid{0000-0002-1717-4133}}

\runningauthor{V. Krug et al.}
\address[A]{Artificial Intelligence Lab, Otto-von-Guericke-University Magdeburg, Germany}
\address[B]{MOTOR Ai GmbH, Berlin, Germany}

\begin{abstract}
Machine Learning with Deep Neural Networks (DNNs) has become a successful tool in solving tasks across various fields of application.
However, the complexity of DNNs makes it difficult to understand how they solve their learned task. 
To improve the explainability of DNNs, we adapt methods from neuroscience that analyze complex and opaque systems. 
Here, we draw inspiration from how neuroscience uses topographic maps to visualize brain activity. 
To also visualize activations of neurons in DNNs as topographic maps, we research techniques to layout the neurons in a two-dimensional space such that neurons of similar activity are in the vicinity of each other. 
In this work, we introduce and compare methods to obtain a topographic layout of neurons in a DNN layer. 
Moreover, we demonstrate how to use topographic activation maps to identify errors or encoded biases and to visualize training processes. 
Our novel visualization technique improves the transparency of DNN-based decision-making systems and is interpretable without expert knowledge in Machine Learning.
\end{abstract}

\begin{keyword}
neural networks \sep deep learning \sep model evaluation \sep explainable AI \sep interpretable AI \sep topographic activation maps
\end{keyword}
\end{frontmatter}
\markboth{May 2023\hb}{May 2023\hb}

\section{Introduction}
Machine Learning with \acp{DNN} is a popular tool and highly successful in solving tasks in many fields of application~\cite{Szegedy2015}. 
However, their complexity complicates the understanding of how they solve their learned task \cite{Yosinski2015}. 
To improve the explainability of \acp{DNN}, we transfer methods from the field of neuroscience, which has been studying the brain over decades. 
In this work, we focus on adapting how brain activity recorded through \ac{EEG} measurements \cite{Makeig2009} is represented as a top view of the head with a superimposed topographic map of neural activity~\cite{maurer2012atlas}. 
We adapt this intuitive visualization of neural activity to use it for \acp{DNN} which do not inherently have a topographic layout.
To this end, we research techniques to layout the neurons in a two-dimensional space in which neurons of similar activity are in the vicinity of each other. 
\acp{SOM} \cite{Kohonen1988} follow a similar motivation and constrain the neurons to form a topographic layout during training. 
However, most \acp{DNN} that are used in practice are trained without such neuron layout.
Our aim is to create a topographic layout visualization for any model, particularly those that are already trained and potentially deployed in the real world.

In this work, we introduce and compare different methods to obtain a topographic layout of neurons in a layer of a \ac{DNN}. 
Moreover, we demonstrate use cases of the resulting visualization. 
This includes identifying potential reasons for erroneous predictions and encoded biases of pre-trained models as well as visualizing training processes.
Our novel visualization technique improves the transparency of \acp{DNN} and is interpretable without expert knowledge in Machine Learning.
In this work, we particularly focus on the visualization of representations. 
Improving models or mitigating biases is out of the scope of this work, but our technique can facilitate applying existing model improvement strategies in a more targeted manner.

\section{Related Work}
Getting insight into the internal structures and processes of trained \acp{DNN} is crucial because such models work as black-boxes \cite{Yosinski2015}.
Consequently, researchers proposed various methods for visualizing and analyzing them.

\paragraph{Feature Visualization}
Feature visualization aims to explain the internal structures of a trained \ac{DL} model.
To investigate which pattern is detected by a convolutional filter, an artificial input which maximally activates this filter can be created \cite{Yosinski2015,Erhan2009,Mordvintsev2015}.
Using a data example or random values as initial input, its values are updated to maximize the activation values of the feature map of interest.
If inputs are optimized without constraints, they can appear unrealistic to a human.
Therefore, the optimization is typically performed with regularization techniques that penalize unnatural inputs \cite{Mordvintsev2015}, which, however, can decrease the faithfulness of the obtained pattern.

\paragraph{Saliency Maps}
To explain the output of a \ac{DNN} for an individual input example, attribution techniques quantify the relevance of each input value for the output \cite{Zeiler2014,Springenberg2015,Kindermans2018,Schulz2019}.
The relevance values are visualized as a heat map on the input, commonly known as a saliency map \cite{Simonyan2013}.
Various attribution techniques have been suggested that, for example, compute gradients \cite{Erhan2009}, combine gradient information with activations \cite{Selvaraju2017} or decompose the output \cite{Bach2015}.
Saliency maps are most suitable for visually interpretable data, for example, images or audio data as spectrograms \cite{Becker2018,Thuillier2018,Perotin2019}.
However, saliency maps only explain individual examples and some attribution techniques can be misleading because their relevance computation is not strongly enough related to the output \cite{Adebayo2018, Nie2018, Sixt2020}.

\paragraph{Data Representation Analysis}
To investigate how the \ac{DNN} processes the input data in general, the model-internal representations of the data can be analyzed.
For example, training linear models to classify the representations in a hidden layer indicates how well particular properties are encoded in this layer \cite{Alain2017,Kim2018}.
The similarity of representations can be used to compare groups of related examples, for example, with \ac{PCA} \cite{Fiacco2019}, \ac{CCA} \cite{Morcos2018a} or clustering techniques \cite{Nagamine2015, krug2021snaps}. 
Further, there are graphical interfaces to investigate representations and learned features \cite{carter2019activation,hohman2019s,park2021neurocartography,wexler2019if}. 
Moreover, the development of representations during training can be studied \cite{li2020visualizing}.
Our introduced technique aims to analyze representations, as well.
In contrast to common existing approaches, it allows to compare high-dimensional representations by visual inspection.

\section{Methods}
\begin{figure}[ht]
	\centering
	\includegraphics[width=\linewidth]{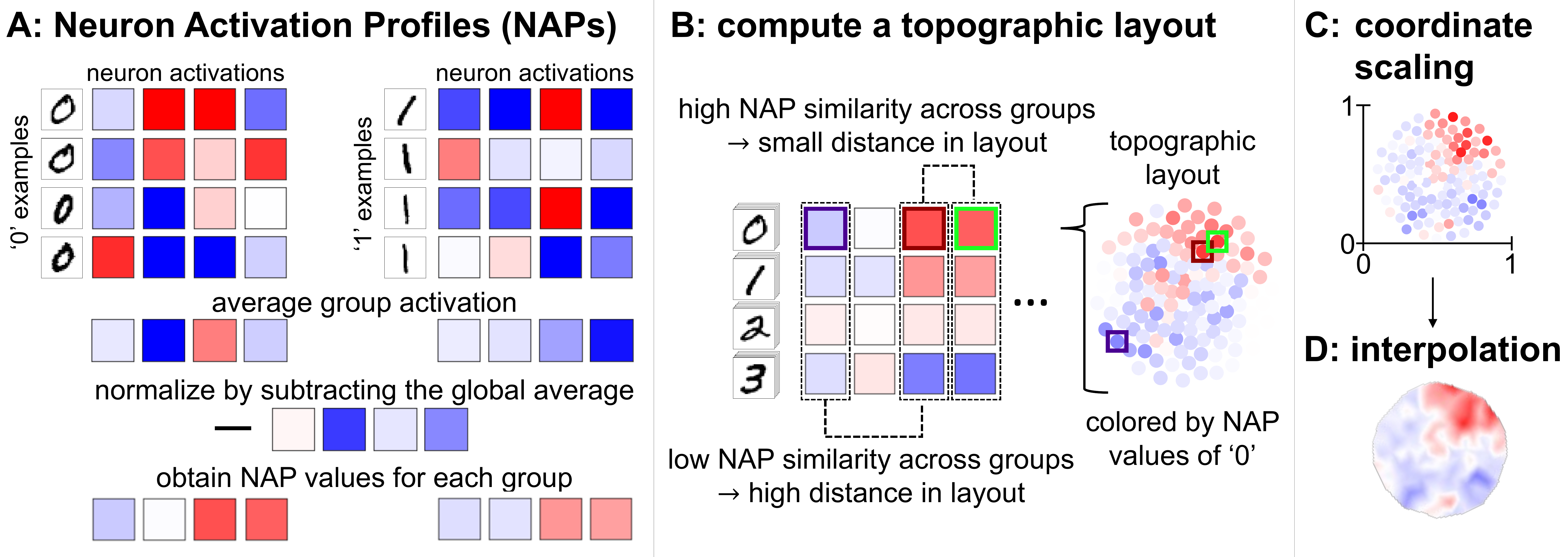}
	\caption{Visual summary of computing topographic activation maps. A: characterize DNN activations for the groups of interest. B: compute a layout in which similarly activated neurons are in the vicinity of each other. C: scale the coordinates in both dimensions to a range from 0 to 1. D: apply interpolation for continuous coloring.}
	\label{fig:methodsummary}
\end{figure}

In this section, we describe our proposed pipeline to compute the topographic activation maps. 
This includes obtaining hidden layer representations of groups of examples, computing the layout of the topographic maps and visualizing activations according to the layout.
A visual abstract of the pipeline is shown in \autoref{fig:methodsummary}.
The implementation is publicly available on \url{https://github.com/valeriekrug/ANN-topomaps}.

\subsection{Hidden Layer Representations}
\label{sec:naps}
We characterize the \ac{DNN} activity with an averaging approach from our previous work \cite{krug2021snaps}.
First, we select groups of interest to compare between, which can be the classes or any other set of groups of examples.
For each group, we compute the average activations in the layer of interest and normalize the result by subtracting the average activation over all groups (\autoref{fig:methodsummary}A).

For the layout computation, we stack the obtained values (\autoref{fig:methodsummary}B) as a $G \times N$ matrix, where $G$ and $N$ denote the number of groups and neurons.
We refer to the resulting matrix as the \ac{NAP} of the layer.

In \acp{CNN}, we characterize the feature map activations instead of each individual neuron.
Therefore, we compute normalized group-averages of the feature maps.
We flatten each to a $w * h$-dimensional vector and concatenate the vectors of all groups ($w * h * G$).
Finally, we obtain the \ac{NAP} of the layer by stacking these vectors for all feature maps, resulting in a $(w * h * G) \times N$ matrix.

\subsection{Topographic Map Layout}
\label{sec:layouting}
To compute the layout of the topographic maps, we distribute the neurons of a hidden layer in a two-dimensional space.
In general, we aim to compute a layout in which neurons of similar activity are in the vicinity of each other (\autoref{fig:methodsummary}B).
In this section, we describe different approaches to obtain such layout.

\subsubsection{Self-Organizing Map (SOM)}
We investigate \acp{SOM} because they are neural networks in which the neurons are arranged in a two-dimensional layout and trained such that neighbors are similar to each other \cite{Kohonen1988}. 
Here, we train a \ac{SOM} on the \ac{NAP} values of the neurons of our investigated model to map them to the topographic layout of the \ac{SOM}.
We use the MiniSom\footnote{https://github.com/JustGlowing/minisom} package to compute a \ac{SOM} layout of the neurons.
For a layer of $N$ neurons, we compute a square \ac{SOM} with shape $d \times d$ with $d=\lfloor \sqrt{N}+1 \rfloor$ such that there can potentially be one neuron per \ac{SOM} position.
We train the \ac{SOM} for 10 epochs and default parameters with the \acp{NAP} as training data. 
Then, we assign each neuron the coordinate (integer values) of the \ac{SOM} position whose weights have the smallest Euclidean distance to its \ac{NAP} values.
To distinguish neurons that are assigned the same coordinate, we distribute them uniformly on a circle (radius of $0.2$) centered at their shared coordinate.

\subsubsection{Co-Activation Graph}
Further, we use a layouting method that considers the most similar pairs of neurons but not their exact similarity values.
To this end, we first compute the pairwise Cosine similarity of neurons according to their \ac{NAP} values.
We then create a graph with nodes representing the neurons.
For the $7.5\%$ most similar pairs of neurons (threshold empirically chosen), we draw an edge between the corresponding nodes.
We further ensure that the entire graph is connected. 
To this end, we identify all connected components and link each smaller connected component to the largest via their most similar pair of neurons.
Finally, we layout the connected graph with the Fruchterman Reingold algorithm \cite{fruchterman1991graph} from the NetworkX package \cite{schult2008exploring} and use the node coordinates as the topographic layout of the neurons.
For brevity, we refer to this technique as the ``graph'' method.

\subsubsection{Dimensionality Reduction}
We test the popular dimensionality reduction methods \acf{PCA} \cite{doi:10.1080/14786440109462720, Hotelling1933AnalysisOA, Jolliffe}, \ac{tSNE} \cite{JMLR:v9:vandermaaten08a} and \ac{UMAP} \cite{mcinnes2020umap}
to project the \ac{NAP} values into a two-dimensional space.
We perform \ac{PCA} using the decomposition module of Scikit-learn (sklearn) \cite{pedregosa2011scikit}, taking the first and second principal components as coordinates.
To compute \ac{tSNE}, we use the sklearn manifold module, initializing the embedding with \ac{PCA}. 
For \ac{UMAP} projection, we use the python module umap-learn\footnote{https://github.com/lmcinnes/umap}. 

\subsubsection{Particle Swarm Optimization (PSO)}
PSO \cite{488968, 699146, 870279} is a biologically inspired algorithm to search for optimal solutions. 
For our approach, we use $N$ particles in a two-dimensional solution space, where each particle corresponds to a neuron. 
Different than the aforementioned approaches, with \ac{PSO} we aim to simultaneously optimize A: clustering of similarly active neurons and B: consistent neuron density.

To achieve activation similarity of neighboring particles, we introduce a global force which is computed based on the \acp{NAP}.
The global force encourages particles of similar neurons to attract each other while repelling dissimilar neurons.
\begin{equation} \label{eq:global_force}
	f_{glob} = attr - rep \quad \textrm{with} \quad attr = a \cdot \left( 1 - \dfrac{dist}{max(dist)^ 3} \right) \quad \textrm{and} \quad rep = b \cdot e^{-(dist/c)}
\end{equation}
In \autoref{eq:global_force}, $f_{glob}$ = global force, $attr$ = global attraction, $rep$ = global repulsion, $dist$~=~Cosine distance matrix of the \acp{NAP}, with $a = 1.5, b = 0.5, c = 2$.

To obtain a well-distributed layout, we use a local force that only depends on the particle coordinates. 
Like the global force, it consists of an attraction and a repulsion term.
In the local force, attraction closes gaps in the layout by penalizing large distances between pairs of particles and repulsion avoids that two particles occupy the same position.
\begin{equation} \label{eq:local_force}
	f_{loc} = attr - rep \quad \textrm{with} \quad attr = a \cdot \left( \dfrac{1}{(dist + 1) ^ 3} \right) \quad \textrm{and} \quad rep = b \cdot e^{-(dist/c)}
\end{equation}
In \autoref{eq:local_force}, $f_{loc}$ = local force, $attr$ = local attraction, $rep$ = local repulsion, $dist$~=~pairwise Euclidean distances between particles, with $a = 1.5, b = 15, c = 2$. 

We optimize the \ac{PSO} for $T=1000$ steps by updating the coordinates according to a weighted average of global and local force (\autoref{eq:combined_force}).
In early steps $t$, we use a high global force weight $w_g$ to encourage the activation similarity of neighboring particles and then gradually increase the local force weight $w_l$ to better distribute the particles.
\begin{equation}
\label{eq:combined_force}
    \begin{gathered}
    	f=\frac{1}{2} \cdot \left( w_g \cdot f_{glob} + w_l \cdot f_{loc} \right) \\ 
     w_l(t) = \frac{1}{2} \cdot \left( \frac{e^{s(t)}-e^{-s(t)}}{e^{s(t)}+e^{-s(t)}}+1 \right) = 1-w_g(t)  \quad\textrm{with }  s(t) = \frac{9 \cdot t}{1000}-3
    \end{gathered}
\end{equation}

\subsubsection{PSO with Non-Random Initialization}
The \ac{PSO} method with random initialization needs careful balancing of the weight parameters of global and local attraction and repulsion.
To require less fine-tuning of parameters, we investigate a variant of the \ac{PSO}. 
We compute an initial similarity-based layout with another layouting method and only use the local force (setting $w_g=0$) to further optimize the resulting layout with \ac{PSO}. 
This way, the \ac{PSO} is only used to equally distribute the neurons in the two-dimensional space.
We call the hybrid methods UMAP\_PSO, TSNE\_PSO, graph\_PSO, SOM\_PSO and PCA\_PSO.

\subsection{Visualization}
Finally, we use the \ac{NAP} values (Section \ref{sec:naps}) and the layout coordinates (Section \ref{sec:layouting}) to create topographic map images.
To be able to compare different layouts, we first scale the layout coordinates to $[0,1]$ in both dimensions (\autoref{fig:methodsummary}C).
Then, we create a mapping from \ac{NAP} values to a symmetric continuous color scale, where blue represents $-max(|NAP|)$, white 0 and red $+max(|NAP|)$. 
We use this mapping to assign colors to the layout coordinates according to the \ac{NAP} values of the respective group.
Then, we linearly interpolate the colors with a resolution of $100 \times 100$~px (\autoref{fig:methodsummary}D).
Equal colors in topographic maps of different groups represent the same \ac{NAP} value, but the colors can correspond to different values in each experiment or layer.
For \acp{CNN}, we color the feature maps by their average \ac{NAP} value.

\section{Experimental Setup}
This section describes the experiments to evaluate the quality of the topographic maps obtained with the proposed layouting methods.
To select the technique with the best quality, we use a simple data set and a shallow model.
In Section~\ref{sec:appl}, we demonstrate that the selected method is also applicable to more complex models and data sets.

\subsection{Data and Models} 
\label{sec:datamodel}
We first test our method with MNIST \cite{lecun-mnisthandwrittendigit-2010}, a common benchmark data set for Machine Learning.
We train a simple \ac{MLP} and \ac{CNN} on the MNIST data set. 
The \ac{MLP} has one fully-connected hidden layer of 128 neurons and uses \ac{ReLU} \cite{nair2010rectified} activation.
The input images are flattened before providing them to the model.
The \ac{CNN} has two 2D-convolutional layers with kernel size 3 $\times$ 3, stride 2 and 128 filters, both using \ac{ReLU} activation.
The fully-connected classification layer takes the flattened feature maps of the second convolutional layer as input.
During training, we use dropout for fully-connected and spatial dropout for convolutional layers, with a dropout rate of 0.5.
Both models are trained using TensorFlow \cite{tensorflow2015-whitepaper}.
We used a batch size of 32 and trained for 20 epochs using the Adam optimizer \cite{Kingma2014} with default parameters and categorical cross-entropy as the loss function.

\subsection{Evaluation Measures}
\paragraph{Qualitative Criteria}
\label{sec:eval:qual}
Our technique aims to provide a comparative visual overview of the representations of groups in a \ac{DNN}.
They shall be perceived as visually similar to topographic maps in neuroscience that have a round shape, contain no empty regions and show continuous sub-regions of similar activity in all groups.
We evaluate these expectations by manual inspection.

\begin{figure}[h]
	\centering
    \includegraphics[width=\linewidth]{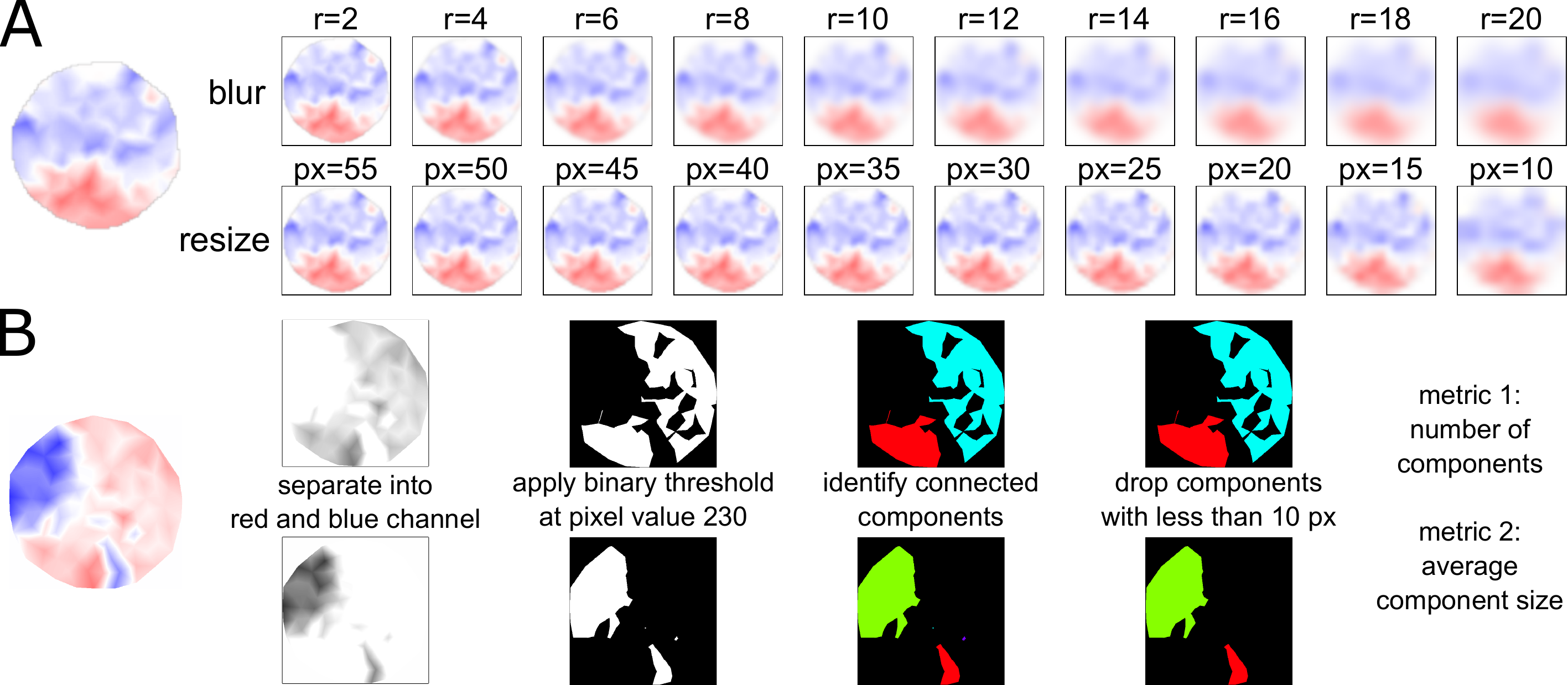}
    \caption{Evaluation metrics for topographic map quality. A: Blur and resize effects used to quantify robustness against perturbations. B: Obtaining quality metrics based on connected components.}
    \label{fig:evalmetrics}
\end{figure}

\paragraph{Quantitative Evaluation Metrics}
\label{sec:eval:quant}
In addition, we quantify the quality of the topographic maps.
To test local activation similarity, we compute the \ac{MSE} between a perturbed version of the image and the original topographic map.
As perturbations, we use Gaussian blur and down- and upscaling with bicubic interpolation.
We use Gaussian blur with radii 2~px to 20~px in steps of 2~px and investigate downscaling sizes to $55 \times 55$~px to $10 \times 10$~px in steps of $5$~px (see~\autoref{fig:evalmetrics}A).
Finally, we aggregate the results for the different parameters with an estimated AUC value (trapezoidal rule).

We further use two metrics that quantify topographic map quality based on connected components in the images.
The procedure can be followed in \autoref{fig:evalmetrics}B.
First, we separate the image of the topographic map into the red and blue channel.
For both channels, we apply a binary threshold at pixel value of 230 to separate blue or red regions from the background.
In the binarized images, we detect connected components using OpenCV\footnote{https://github.com/opencv/opencv-python}.
We omit components smaller than 10~px area because they are not perceived as distinct regions.
Finally, we compute the number of components and their average size.
Generally, few large components are considered as high quality.
However, a single large component is uninformative because there are no discriminable regions.

As the quality differs between groups, we report the average quality over the groups.
Further, we investigate the robustness of the quality of the topographic maps.
To this end, we repeat each topographic map computation 100 times given the same input.

\subsection{Experimental Plan}
\label{sec:exp}
For our simplest data set and model, MNIST and MLP, we compute \acp{NAP} in the first fully-connected layer, using the 10 classes as grouping. 
We then use the resulting \acp{NAP} to compute topographic maps with each of our 11 proposed layouting methods.

First, we pre-select techniques that satisfy the qualitative expectations of the visualization described in Section~\ref{sec:eval:qual}.
For an exemplary coloring of the layout for the class ``0'', we compare the methods with respect to the formation of regions of similar activations and the visual similarity to a topographic map in neuroscience.
Next, for the pre-selected methods, we compare the quality of the topographic maps according to the quantitative measures described in Section~\ref{sec:eval:quant}. 
For \acp{CNN}, we use \acp{NAP} of the feature maps to compute the topographic map layouts but average the activation values per feature map to obtain the colors.

\section{Results and Discussion}
\label{sec:results}

\subsection{Pre-selecting Layouting Methods}
\label{sec:res:qual}
\begin{figure}[t]
	\centering
	\includegraphics[width=0.90\linewidth]{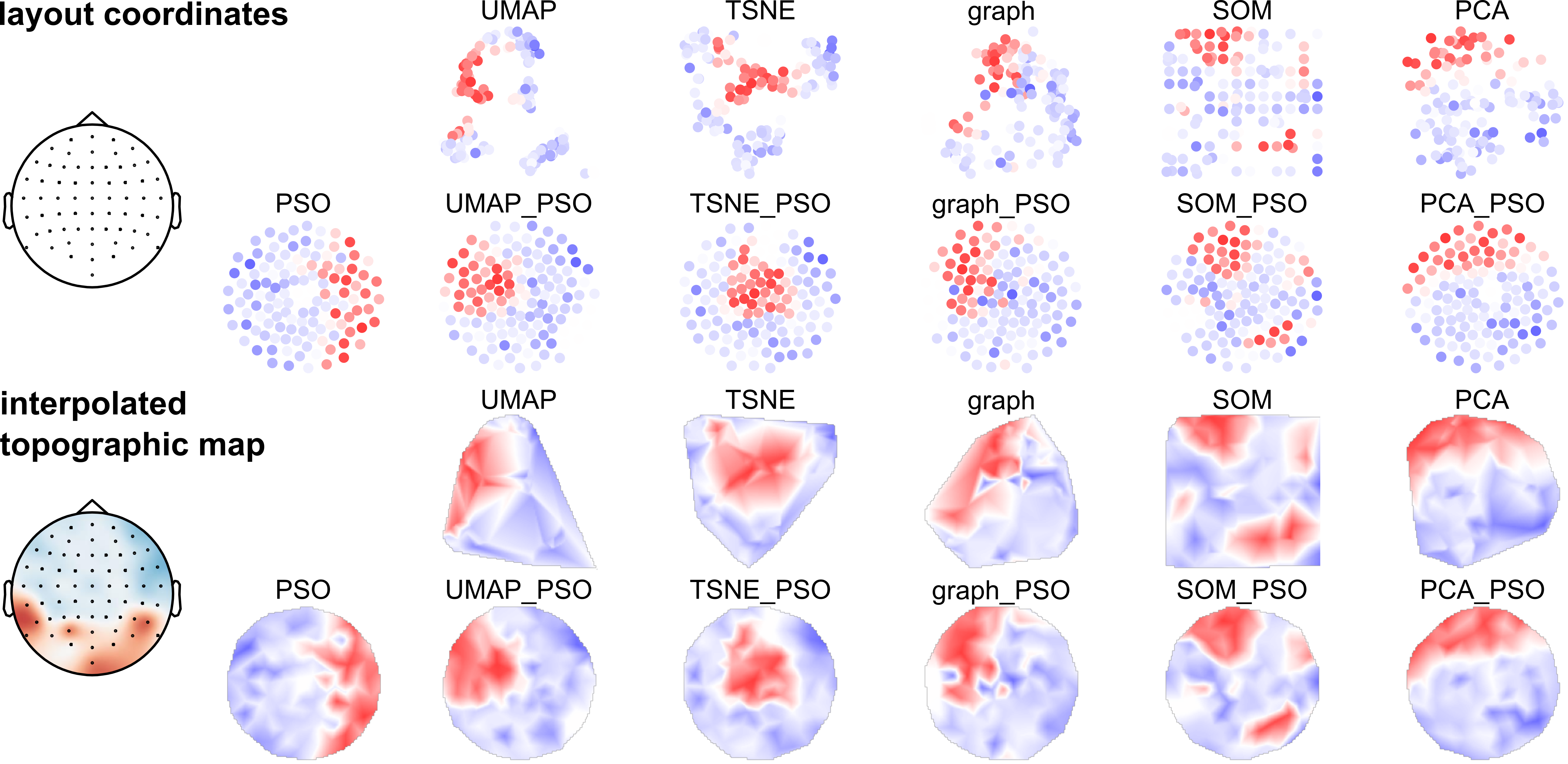}
	\caption{Topographic maps for one exemplary class for all proposed layouting methods. The scatter plots show the layouted neurons with \ac{NAP} value-based coloring. Below are the resulting interpolated topographic maps. Left-most are examples of an electrode layout (top) and a topographic map (bottom) in neuroscience. All layouts and colorings use the same class-based \acp{NAP} for an MNIST \ac{MLP} model as input. 
	}
	\label{fig:layouteval}
\end{figure}
We first pre-select layouting methods that produce topographic maps which satisfy the qualitative criteria of Section~\ref{sec:eval:qual}.
Topographic maps generated for the MNIST \ac{MLP} and colored by the exemplary class ``0'' are shown in \autoref{fig:layouteval}.
All methods distribute the neurons to form regions of similar activations.
Only the \ac{SOM} technique splits up sets of co-activated neurons into multiple regions in the layout due to not penalizing similarity of distant coordinates.
Another criterion is that the neurons are well-distributed in the two-dimensional space.
This is crucial because a layout with varying neuron density leads to disproportionate regions in the interpolated topographic maps.
For example, in the interpolated TSNE topographic map, the gaps cause the red region to be enlarged, which wrongly suggests that the highly active neurons are in the majority.
As expected, the best distribution is achieved with the PSO methods due to their local force component.
Further, we favor topographic maps of round shape.
This property is best achieved with the \ac{PSO} methods due to the local force.
This supports our idea of first layouting the neurons by activation similarity and then distributing the neurons with a \ac{PSO} using the local force only.
Therefore, we conclude that the PSO methods are the most promising techniques and use them for the quantitative evaluation.
As a baseline, we include a PSO that only uses local force, initialized with random uniform coordinates.

\subsection{Quantitative Evaluation}
\label{sec:res:mlps}
\begin{figure}[h]
	\centering
	\includegraphics[width=\linewidth]{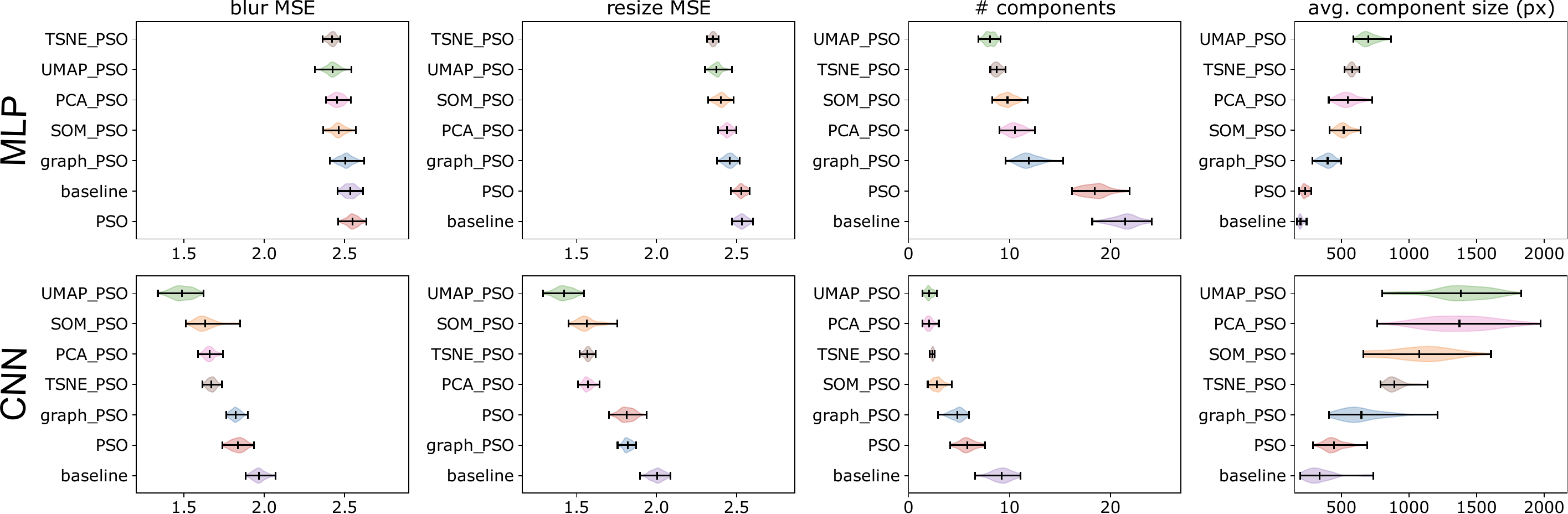}
	\caption{Quantification of the topographic map quality (average of the groups) for different metrics across 100 trials, highlighting the mean value and the extrema with markers. The rows of each violin plot are sorted by the respective mean values, decreasing in quality from the top to the bottom row.}
	\label{fig:topoeval}
\end{figure}

For the pre-selected methods, we quantify the topographic map quality.
Generally, we consider low MSE values and few components of large size as high quality.
However, a topographic map that only shows a single large component is uninformative because there are no discriminable regions.

The results for the first fully-connected layer of the \ac{MLP} and the first convolutional layer of the CNN trained on MNIST are shown in \autoref{fig:topoeval}.
In the \ac{MLP} (\autoref{fig:topoeval} top), the lowest quality is shared by the baseline and the PSO layout.
graph\_PSO, SOM\_PSO and PCA\_PSO are in the medium quality range, while graph\_PSO shows the lowest quality among the three methods.
UMAP\_PSO and TSNE\_PSO obtain the highest quality results for the \ac{MLP} according to all metrics.
TSNE\_PSO shows smaller MSE values than UMAP\_PSO, but not significantly.
Regarding the component metrics, UMAP\_PSO has a clearly higher quality than TSNE\_PSO.
The high quality of the two methods also comes with longer computation time.
This is negligible for computing one layout but accumulates when visualizing several layers in a deep network.

In the \ac{CNN} (\autoref{fig:topoeval} bottom), UMAP\_PSO shows the best rank according to all quality metrics.
Different to our findings for \acp{MLP}, TSNE\_PSO is not as high-quality as UMAP\_PSO anymore and is lower-quality than SOM\_PSO or PCA\_PSO in many cases.
PSO with random initialization leads to low-quality topographic maps like in the \ac{MLP}, but graph\_PSO yields better results than the default PSO .
We conclude that UMAP\_PSO produces highest-quality topographic maps for both \ac{MLP} and \ac{CNN} and use it in the following applications.

\section{Exemplary Applications}
\label{sec:appl}

\subsection{Detecting Systematic Annotation Errors in Data Sets}
Topographic map visualizations can be used to identify whether classification errors are caused by wrong annotations that can can occur in training data or test data.
We demonstrate how to use our visualization technique for two toy examples, where we introduce annotation errors in either the training or test data.

\paragraph{Toy Examples Design} 
For the first toy example, we use the Fashion MNIST data set \cite{Xiao2017}.
In the test data, we introduce a systematic error by changing 
the target class of 90\% of the examples of class ``T-shirt/Top'' to class ``trouser''.
Using this altered data, we create topographic maps for a \ac{MLP} model, trained on the original training data.

In the second toy example, we use the MNIST data set and change 90\% of the class ``0'' examples to class ``1'', train a \ac{MLP} model on this altered training data set and create topographic maps for the original MNIST test data.

Both toy examples use the shallow \ac{MLP} architecture as described in Section~\ref{sec:datamodel}.
The 20 groups of interest are the test data labels separated into whether they are correctly predicted.
We create topographic maps for the first fully-connected layer of the respective \ac{MLP} model using \ac{NAP} values computed for 200 random examples per group.

\begin{figure}[t]
	\centering
	\includegraphics[width=\linewidth]{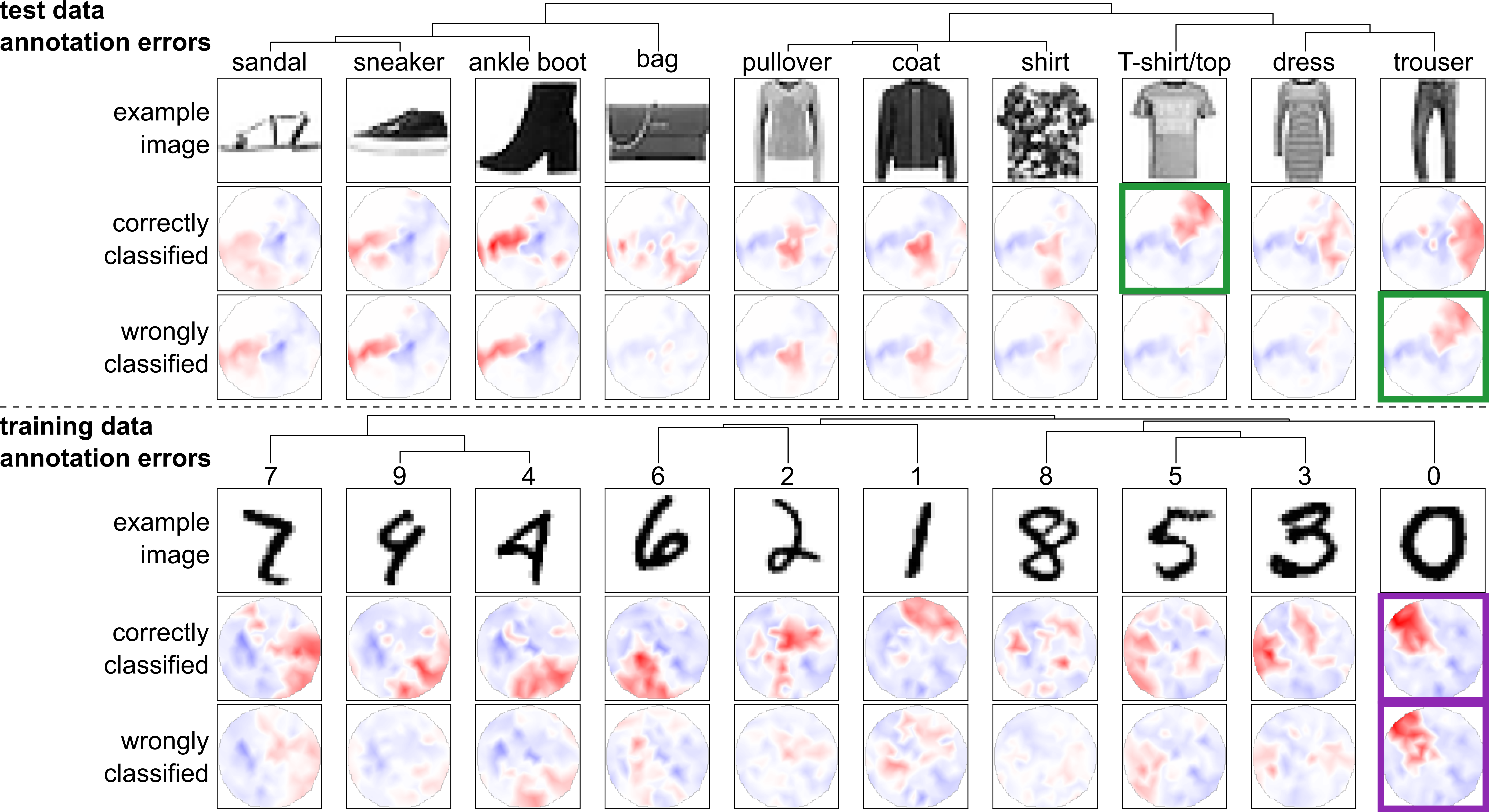}
	\caption{Topographic maps for correctly and wrongly classified examples using data sets with annotation errors. Top: annotation errors in the test data. Bottom: training data with annotation errors. The activation similarity is shown as a dendrogram and used to sort the classes. The shown example images are randomly chosen from the respective group, while 200 examples per group are used to compute the \ac{NAP}. The green and purple annotations highlight the pairs of topographic maps that indicate the error in the respective example.}
	\label{fig:annerrors}
\end{figure}

\paragraph{Annotation Errors in the Test Data}
\autoref{fig:annerrors} (top) shows the topographic maps of all Fashion MNIST classes.
In the shown example, there is a high activation difference between correctly and wrongly classified examples for several classes, for example, ``bag'' and ``trouser''.
In realistic models, such dissimilarity indicates a distribution difference between training and test data for these classes.
Here, for the wrongly classified ``trouser'' group, we further observe a high similarity to the activation of correctly classified ``T-shirt/top'' images (highlighted in green), indicating the annotation error we introduced.

Notably, in the upper left of the topographic maps, we observe a white region in all groups.
Such region corresponds to a subset of neurons whose activity does not differ between groups.
This indicates that the model does not use its full capacity because it is overcomplex or due to training problems like \acp{ReLU} that never activate \cite{maas2013rectifier}.

\paragraph{Annotation Errors in the Training Data}
\label{sec:appl:trainerrors}
To investigate training data annotation errors, we use a model trained on an erroneous MNIST data set.
The resulting topographic maps using the original MNIST data set are shown in \autoref{fig:annerrors} (bottom).
Annotation errors in the training data lead to highly similar activations for wrongly and correctly classified examples of the same group.
We observe this pattern for class ``0'' (purple highlight), which again matches our injected annotation error of changing 90\% of the ``0'' labels.

This pattern also occurs if the model cannot discriminate between two or more classes properly, for example, in the classes ``sneaker'' and ``ankle boot'' of \autoref{fig:annerrors} (top).

\subsection{Visualization of Bias in Representations}
\label{sec:appl:bias}
In this section, we demonstrate how to detect racial bias with our technique.
Different to common approaches that investigate bias in the model output \cite{pmlr-v81-buolamwini18a,karkkainenfairface}, we focus on bias in the representations of a \ac{DNN}.
Here, we inspect VGG16 \cite{DBLP:journals/corr/SimonyanZ14a}, which can be used as feature extractor for downstream tasks like image recognition.
As test data, we use FairFace \cite{karkkainenfairface}, a balanced data set of images of people from different age groups, races and binary genders.
We choose the ``race'' variable as grouping to compute the topographic maps.
As random baseline, we add groups of random examples.
We obtained VGG16 from TensorFlow Keras applications\footnote{https://github.com/keras-team/keras}, using the second maxpooling layer as an example.

\begin{figure}[h]
	\centering
	\includegraphics[width=\linewidth]{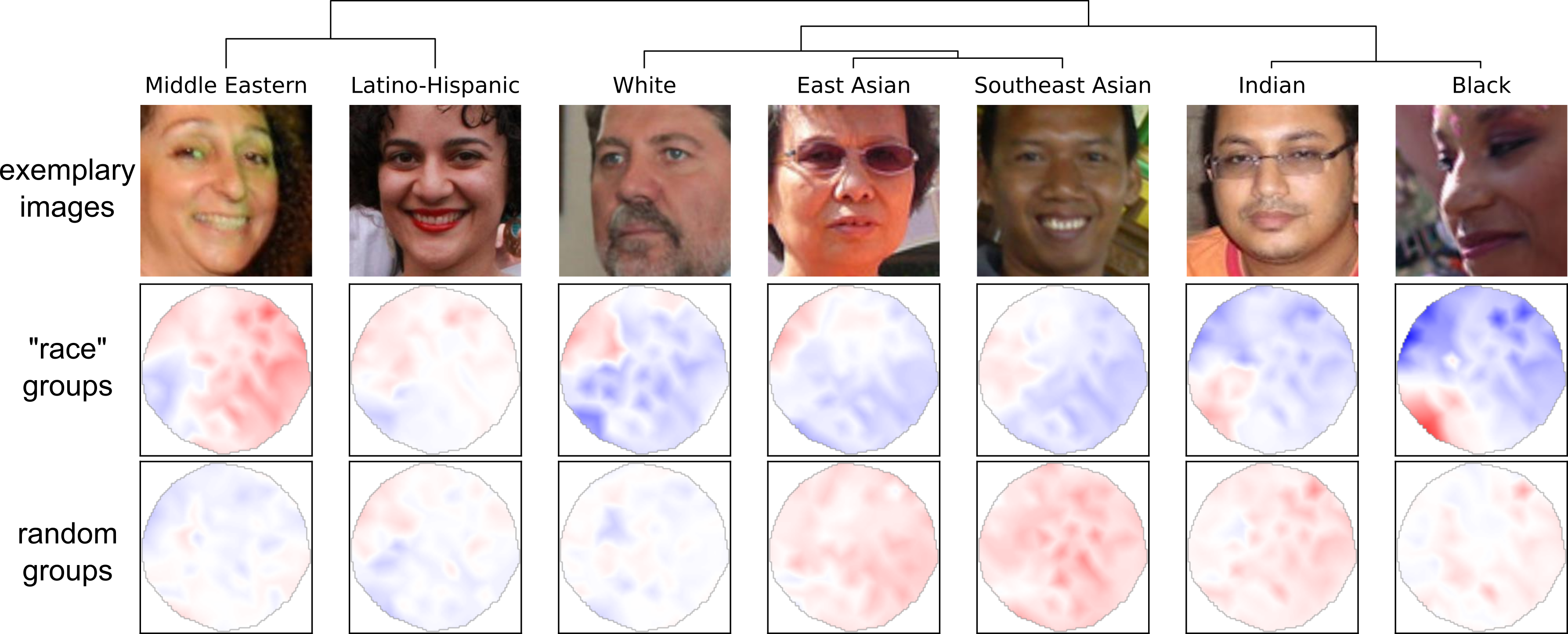}
	\caption{Topographic maps of VGG16 activations in the second maxpooling layer for different FairFace ``race'' categories (middle) and random groups (bottom). In each row, groups are sorted by activation similarity.}
	\label{fig:bias_l6}
\end{figure}

By comparing the topographic maps between the values of the sensitive variables from FairFace, we investigate whether it is possible to visually discriminate the group representations of the VGG16 model.
\autoref{fig:bias_l6} shows topographic maps for the seven ``race'' categories of the FairFace data set and of seven random groups in the second maxpooling layer of the VGG16 model.
First, we observe that it is clearly easier to discriminate the ``race'' categories than the random groups, indicating a racial bias.
Only the ``Latino-Hispanic'' topographic map can be confused with a random group.
We further observe that ``Indian'' and ``Black'' are perceived as particularly similar by the model. 
``East Asian`` and ``Southeast Asian`` are similar to each other and to the ``White'' category.
``Middle Eastern'' and ``Latino-Hispanic'' are dissimilar to the other groups.

Observing visually discriminable representations indicates that downstream applications which use the pre-trained model can reproduce the bias.
For example, a classifier can easily learn different decisions for the ``White'' and ``Black'' categories.
We emphasize that this observation does not imply that a downstream application must include a racial bias.
Instead, we suggest to use the findings to formulate hypotheses about which bias to look for.
This allows to test for likely biases in a targeted way.  

\subsection{Visualizing Training Processes}
In the third exemplary application, we show how the topographic maps can reveal information about the training process of a \ac{DNN}. 
For this, we train a simple \ac{MLP} model on MNIST for one epoch and saved the model after each batch resulting in a total of 1875 model states. 
For the last state, that is, the fully-trained model, we compute a topographic map layout using UMAP\_PSO and use it for all other points in training.
This way, the layout stays the same and we observe the development of activations during the training.

\begin{figure}[t]
	\centering
	\includegraphics[width=\linewidth]{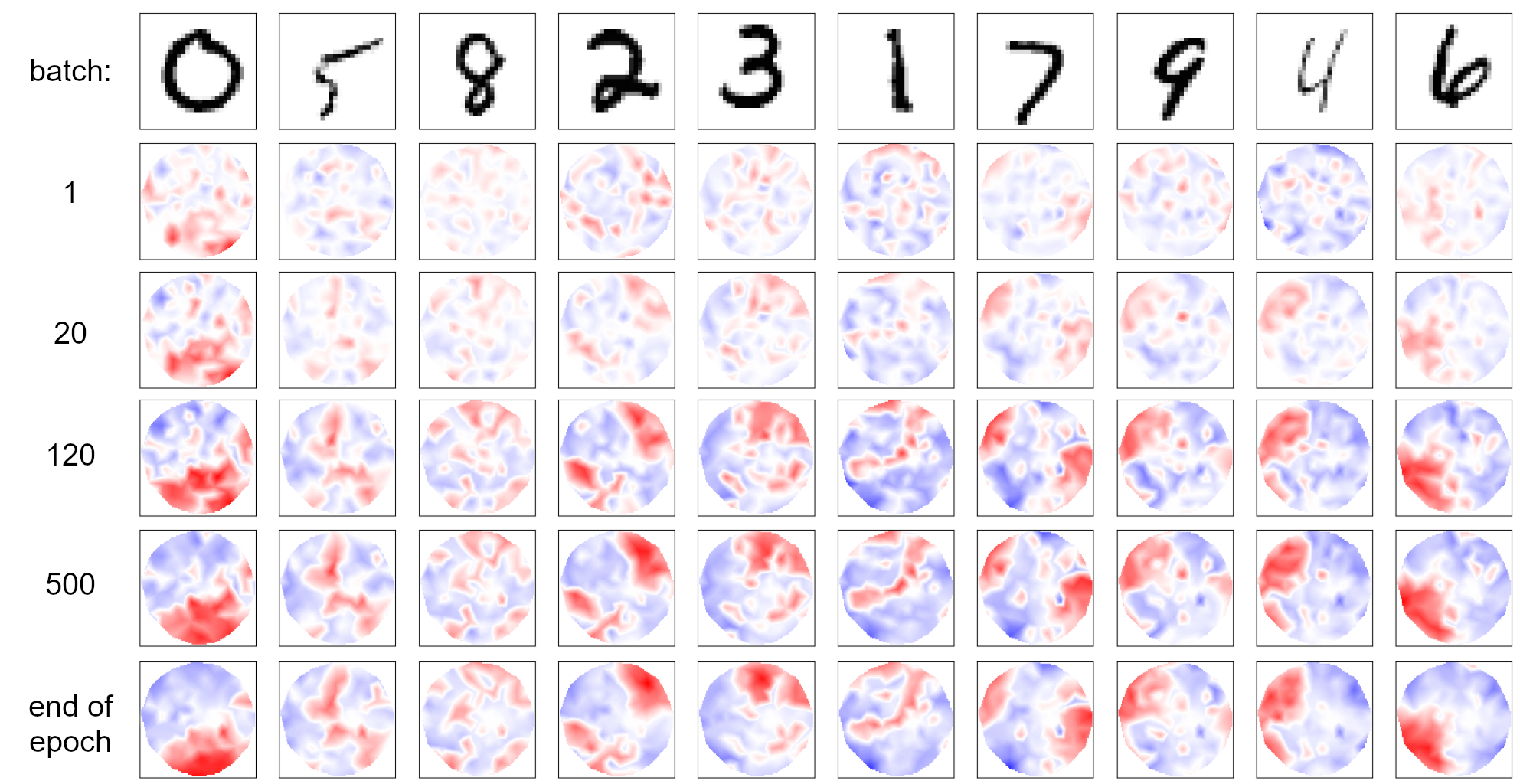}
	\caption{Topographic maps at different stages of training. Classes are ordered by activation similarity.}
	\label{fig:topo_training}
\end{figure}

\autoref{fig:topo_training} shows a training process as topographic maps at exemplary time steps. 
We observe, that the change of \ac{DNN} activations is particularly large in the first 100 batches and gradually changes less as training continues. 
Correspondingly, we show a smaller batch difference early in training than towards the end of training.
We observe that initial shapes begin to emerge after approximately 20 batches. 
These shapes become more sophisticated after around 120 batches and continue to become clearer over time. 
After around 500 batches, we only observe minimal changes.
Further, we observe that the topographic maps of some classes stabilize earlier than others, which indicates that they are easier to learn for the model.
For example, classes ``1'' and ``4'' change little from batch 500 to the end of the epoch, while some active regions disappear for classes ``0'' and ``6''.
Li et al. \cite{li2020visualizing} found different classes to be learned early and late by their used \ac{CNN} model, indicating that these learning dynamics are model-dependent. 
Topographic maps provide a unique perspective on the training process and help to better understand the dynamics of the neural network training.
However, due to saving and processing many model states, the analysis of training processes requires more computation and storage resources than for a trained model only.

\section{Conclusion} 
Topographic activation maps are a promising tool to get insight into the internal representations of \acp{DNN}.
Our technique simplifies hidden layer activations as a two-dimensional visualization to provide a comparative overview of representations of groups of inputs.
It is capable of showing activations of each neuron in \acp{MLP} but \ac{CNN} feature maps need to be aggregated such that large feature maps might not be represented optimally.

Our technique alone does not provide explanations of the model representations or decisions.
It still requires a human to interpret the results or to perform further downstream analyses to explain what the regions are responsible for.
While topographic maps are easy to interpret by visual inspection, relating the visualization to useful insight into the model requires practice, especially for highly-complex models that are used in practical applications.
We therefore recommend to first get familiar with the technique by using toy examples before applying it to real-world models.

In future research, we will investigate how to generate explanations of the regions in our topographic map visualization.
For example, we will automatically detect group-responsive regions and perform feature visualization for the corresponding filters of the model to understand which patterns it uses to detect the group.
Moreover, we will extend our technique to create multi-layer topographic maps.

\bibliographystyle{vancouver}
\bibliography{references} 

\begin{thebibliography}{10}

\bibitem{Szegedy2015}
Szegedy C, Liu W, Jia Y, Sermanet P, Reed S, Anguelov D, et~al.
\newblock Going deeper with convolutions.
\newblock In: IEEE Conference on Computer Vision and Pattern Recognition
  (CVPR); 2015. p. 1-9.

\bibitem{Yosinski2015}
Yosinski J, Clune J, Nguyen A, Fuchs T, Lipson H.
\newblock Understanding Neural Networks Through Deep Visualization.
\newblock arXiv preprint arXiv:150606579. 2015.

\bibitem{Makeig2009}
Makeig S, Onton J, et~al.
\newblock {ERP} features and {EEG} dynamics: an {ICA} perspective.
\newblock In: Oxford handbook of event-related potential components. Oxford;
  2009. p. 51-87.

\bibitem{maurer2012atlas}
Maurer K, Dierks T.
\newblock Atlas of Brain Mapping: Topographic Mapping of EEG and Evoked
  Potentials.
\newblock Springer Science \& Business Media; 2012.

\bibitem{Kohonen1988}
Kohonen T.
\newblock In: Self-Organizing Feature Maps. Berlin, Heidelberg: Springer Berlin
  Heidelberg; 1988. p. 119-57.

\bibitem{Erhan2009}
Erhan D, Bengio Y, Courville A, Vincent P.
\newblock Visualizing higher-layer features of a deep network.
\newblock University of Montreal. 2009;1341(3):1.

\bibitem{Mordvintsev2015}
Mordvintsev A, Olah C, Tyka M.
\newblock Inceptionism: Going deeper into neural networks.
\newblock Google Research Blog Retrieved June. 2015;20(14):5.

\bibitem{Zeiler2014}
Zeiler MD, Fergus R.
\newblock Visualizing and Understanding Convolutional Networks.
\newblock In: European Conference on Computer Vision (ECCV). Springer; 2014. p.
  818-33.

\bibitem{Springenberg2015}
Springenberg JT, Dosovitskiy A, Brox T, Riedmiller M.
\newblock Striving for simplicity: The all convolutional net.
\newblock arXiv preprint arXiv:14126806. 2014.

\bibitem{Kindermans2018}
Kindermans PJ, Schütt KT, Alber M, Müller KR, Erhan D, Kim B, et~al.
\newblock Learning how to explain neural networks: PatternNet and
  PatternAttribution.
\newblock International Conference on Learning Representations (ICLR). 2018.

\bibitem{Schulz2019}
Schulz K, Sixt L, Tombari F, Landgraf T.
\newblock Restricting the Flow: Information Bottlenecks for Attribution.
\newblock International Conference on Learning Representations (ICLR). 2019.

\bibitem{Simonyan2013}
Simonyan K, Vedaldi A, Zisserman A.
\newblock Deep inside convolutional networks: Visualising image classification
  models and saliency maps.
\newblock arXiv preprint arXiv:13126034. 2013.

\bibitem{Selvaraju2017}
Selvaraju RR, Cogswell M, Das A, Vedantam R, Parikh D, Batra D.
\newblock Grad-CAM: Visual Explanations From Deep Networks via Gradient-Based
  Localization.
\newblock In: IEEE International Conference on Computer Vision (ICCV); 2017. p.
  618-26.

\bibitem{Bach2015}
Bach S, Binder A, Montavon G, Klauschen F, M{\"u}ller KR, Samek W.
\newblock On pixel-wise explanations for non-linear classifier decisions by
  layer-wise relevance propagation.
\newblock PloS one. 2015;10(7):e0130140.

\bibitem{Becker2018}
Becker S, Ackermann M, Lapuschkin S, M{\"u}ller KR, Samek W.
\newblock Interpreting and explaining deep neural networks for classification
  of audio signals.
\newblock arXiv preprint arXiv:180703418. 2018.

\bibitem{Thuillier2018}
Thuillier E, Gamper H, Tashev IJ.
\newblock Spatial audio feature discovery with convolutional neural networks.
\newblock In: IEEE International Conference on Acoustics, Speech and Signal
  Processing (ICASSP); 2018. p. 6797-801.

\bibitem{Perotin2019}
Perotin L, Serizel R, Vincent E, Gu{\'e}rin A.
\newblock CRNN-based multiple DoA estimation using acoustic intensity features
  for Ambisonics recordings.
\newblock IEEE Journal of Selected Topics in Signal Processing.
  2019;13(1):22-33.

\bibitem{Adebayo2018}
Adebayo J, Gilmer J, Muelly M, Goodfellow I, Hardt M, Kim B.
\newblock Sanity checks for saliency maps.
\newblock In: Advances in Neural Information Processing Systems; 2018. p.
  9525-36.

\bibitem{Nie2018}
Nie W, Zhang Y, Patel A.
\newblock A theoretical explanation for perplexing behaviors of
  backpropagation-based visualizations.
\newblock In: International Conference on Machine Learning (ICML); 2018. p.
  3809-18.

\bibitem{Sixt2020}
Sixt L, Granz M, Landgraf T.
\newblock When Explanations Lie: Why Many Modified BP Attributions Fail.
\newblock In: International Conference on Machine Learning (ICML); 2020. p.
  9046-57.

\bibitem{Alain2017}
Alain G, Bengio Y.
\newblock Understanding intermediate layers using linear classifier probes.
\newblock International Conference on Learning Representations (ICLR), Workshop
  Track Proceedings. 2017.

\bibitem{Kim2018}
Kim B, Wattenberg M, Gilmer J, Cai C, Wexler J, Viegas F, et~al.
\newblock Interpretability Beyond Feature Attribution: Quantitative Testing
  with Concept Activation Vectors (TCAV).
\newblock In: International Conference on Machine Learning (ICML); 2018. p.
  2668-77.

\bibitem{Fiacco2019}
Fiacco J, Choudhary S, Rose C.
\newblock Deep neural model inspection and comparison via functional neuron
  pathways.
\newblock In: Annual Meeting of the Association for Computational Linguistics
  (ACL); 2019. p. 5754-64.

\bibitem{Morcos2018a}
Morcos AS, Raghu M, Bengio S.
\newblock Insights on representational similarity in neural networks with
  canonical correlation.
\newblock arXiv preprint arXiv:180605759. 2018.

\bibitem{Nagamine2015}
Nagamine T, Seltzer ML, Mesgarani N.
\newblock Exploring how deep neural networks form phonemic categories.
\newblock Sixteenth Annual Conference of the International Speech Communication
  Association. 2015.

\bibitem{krug2021snaps}
Krug A, Ebrahimzadeh M, Alemann J, Johannsmeier J, Stober S.
\newblock Analyzing and Visualizing Deep Neural Networks for Speech Recognition
  with Saliency-Adjusted Neuron Activation Profiles.
\newblock MDPI Electronics. 2021;10(11):1350.

\bibitem{carter2019activation}
Carter S, Armstrong Z, Schubert L, Johnson I, Olah C.
\newblock Activation atlas.
\newblock Distill. 2019;4(3):e15.

\bibitem{hohman2019s}
Hohman F, Park H, Robinson C, Chau DHP.
\newblock S ummit: Scaling deep learning interpretability by visualizing
  activation and attribution summarizations.
\newblock IEEE transactions on visualization and computer graphics.
  2019;26(1):1096-106.

\bibitem{park2021neurocartography}
Park H, Das N, Duggal R, Wright AP, Shaikh O, Hohman F, et~al.
\newblock Neurocartography: Scalable automatic visual summarization of concepts
  in deep neural networks.
\newblock IEEE Transactions on Visualization and Computer Graphics.
  2021;28(1):813-23.

\bibitem{wexler2019if}
Wexler J, Pushkarna M, Bolukbasi T, Wattenberg M, Vi{\'e}gas F, Wilson J.
\newblock The what-if tool: Interactive probing of machine learning models.
\newblock IEEE transactions on visualization and computer graphics.
  2019;26(1):56-65.

\bibitem{li2020visualizing}
Li M, Zhao Z, Scheidegger C.
\newblock Visualizing neural networks with the grand tour.
\newblock Distill. 2020;5(3):e25.

\bibitem{fruchterman1991graph}
Fruchterman TM, Reingold EM.
\newblock Graph Drawing by Force-Directed Placement.
\newblock Software: Practice and experience. 1991;21:1129-64.

\bibitem{schult2008exploring}
Schult DA; Citeseer.
\newblock Exploring network structure, dynamics, and function using NetworkX.
\newblock In Proceedings of the 7th Python in Science Conference (SciPy). 2008.

\bibitem{doi:10.1080/14786440109462720}
S KPFR.
\newblock On lines and planes of closest fit to systems of points in space.
\newblock The London, Edinburgh, and Dublin Philosophical Magazine and Journal
  of Science. 1901;2(11):559-72.

\bibitem{Hotelling1933AnalysisOA}
Hotelling H.
\newblock Analysis of a Complex of Statistical Variables into Principal
  Components.
\newblock Journal of Educational Psychology. 1933;24:498-520.

\bibitem{Jolliffe}
{Jolliffe} I.
\newblock Principal Component Analysis.
\newblock New York: Springer Verlag; 2002.

\bibitem{JMLR:v9:vandermaaten08a}
van~der Maaten L, Hinton G.
\newblock Visualizing Data using t-SNE.
\newblock Journal of Machine Learning Research. 2008;9(86):2579-605.

\bibitem{mcinnes2020umap}
McInnes L, Healy J, Melville J.
\newblock UMAP: Uniform Manifold Approximation and Projection for Dimension
  Reduction.
\newblock arXiv preprint arXiv:180203426. 2020.

\bibitem{pedregosa2011scikit}
Pedregosa F, Varoquaux G, Gramfort A, Michel V, Thirion B, Grisel O, et~al.
\newblock Scikit-learn: Machine learning in Python.
\newblock The Journal of Machine Learning Research. 2011;12:2825-30.

\bibitem{488968}
Kennedy J, Eberhart R.
\newblock Particle Swarm Optimization.
\newblock In: Proceedings of International Conference on Neural Networks
  (ICCN). vol.~4; 1995. p. 1942-8 vol.4.

\bibitem{699146}
Shi Y, Eberhart R.
\newblock A Modified Particle Swarm Optimizer.
\newblock In: IEEE International Conference on Evolutionary Computation
  Proceedings.; 1998. p. 69-73.

\bibitem{870279}
Eberhart RC, Shi Y.
\newblock Comparing Inertia Weights and Constriction Factors in Particle Swarm
  Optimization.
\newblock In: Proceedings of the Congress on Evolutionary Computation.. vol.~1;
  2000. p. 84-8 vol.1.

\bibitem{lecun-mnisthandwrittendigit-2010}
LeCun Y, Cortes C. {MNIST} handwritten digit database; 2010.
\newblock http://yann.lecun.com/exdb/mnist/.

\bibitem{nair2010rectified}
Nair V, Hinton GE.
\newblock Rectified Linear Units Improve Restricted Boltzmann Machines.
\newblock In: International Conference on Machine Learning (ICML); 2010. p.
  807-14.

\bibitem{tensorflow2015-whitepaper}
Abadi M, Agarwal A, Barham P, Brevdo E, Chen Z, Citro C, et~al.. {TensorFlow}:
  Large-Scale Machine Learning on Heterogeneous Systems; 2015.
\newblock Software available from tensorflow.org.
\newblock Available from: \url{https://www.tensorflow.org/}.

\bibitem{Kingma2014}
Kingma DP, Ba J.
\newblock Adam: A method for stochastic optimization.
\newblock arXiv preprint arXiv:14126980. 2014.

\bibitem{Xiao2017}
Xiao H, Rasul K, Vollgraf R.
\newblock Fashion-{MNIST}: a novel image dataset for benchmarking machine
  learning algorithms.
\newblock arXiv preprint arXiv:170807747. 2017.

\bibitem{maas2013rectifier}
Maas AL, Hannun AY, Ng AY, et~al.
\newblock Rectifier Nonlinearities Improve Neural Network Acoustic Models.
\newblock In: International Conference on Machine Learning (ICML). vol.~30.
  Citeseer; 2013. p.~3.

\bibitem{pmlr-v81-buolamwini18a}
Buolamwini J, Gebru T.
\newblock Gender Shades: Intersectional Accuracy Disparities in Commercial
  Gender Classification.
\newblock In: Friedler SA, Wilson C, editors. Proceedings of the 1st Conference
  on Fairness, Accountability and Transparency. vol.~81. PMLR; 2018. p. 77-91.

\bibitem{karkkainenfairface}
Karkkainen K, Joo J.
\newblock FairFace: Face Attribute Dataset for Balanced Race, Gender, and Age
  for Bias Measurement and Mitigation.
\newblock In: Proceedings of the IEEE/CVF Winter Conference on Applications of
  Computer Vision; 2021. p. 1548-58.

\bibitem{DBLP:journals/corr/SimonyanZ14a}
Simonyan K, Zisserman A.
\newblock Very deep convolutional networks for large-scale image recognition.
\newblock arXiv preprint arXiv:14091556. 2014.

\end{thebibliography}
\end{document}